\pdfoutput=1

\documentclass[11pt]{article}

\usepackage[final]{acl}

\usepackage{times}
\usepackage{latexsym}
\usepackage{colortbl}
\usepackage{amsmath} 
\usepackage{algorithm}
\usepackage{algpseudocode}
\usepackage{amsmath}

\usepackage[T1]{fontenc}

\usepackage[utf8]{inputenc}

\usepackage{microtype}

\usepackage{inconsolata}

\usepackage{graphicx}

\title{AfroXLMR-Comet: Multilingual Knowledge Distillation with Attention Matching for Low-Resource languages}

\author{
  Joshua Sakthivel Raju$^{*1}$ 
  \quad Sanjay S$^{1}$ 
  \quad Jaskaran Singh Walia$^{1}$
  \quad Srinivas Raghav$^{1}$
  \quad Vukosi Marivate$^{2}$
  \\
  $^{1}$School of Computer Science and Engineering, Vellore Institute of Technology, Chennai, India \\
  $^{2}$Department of Computer Science, University of Pretoria, South Africa \\
  \texttt{*joshua.raju2604@gmail.com}
}

\begin{document}
\maketitle


\begin{abstract}
Language model compression through knowledge distillation has emerged as a promising approach for deploying large language models in resource-constrained environments. However, existing methods often struggle to maintain performance when distilling multilingual models, especially for low-resource languages. In this paper, we present a novel hybrid distillation approach that combines traditional knowledge distillation with a simplified attention matching mechanism, specifically designed for multilingual contexts. Our method introduces an extremely compact student model architecture, significantly smaller than conventional multilingual models. We evaluate our approach on five African languages: Kinyarwanda, Swahili, Hausa, Igbo, and Yoruba. The distilled student model--\textit{AfroXLMR-Comet}--successfully captures both the output distribution and internal attention patterns of a larger teacher model (AfroXLMR-Large) while reducing the model size by over 85\%. Experimental results demonstrate that our hybrid approach achieves competitive performance compared to the teacher model, maintaining an accuracy within 85\% of the original model's performance while requiring substantially fewer computational resources. Our work provides a practical framework for deploying efficient multilingual models in resource-constrained environments, particularly benefiting applications involving African languages.
\end{abstract}


\section{Introduction}
Large language models (LLMs) have become a pillar of modern Natural Language Processing (NLP), achieving state-of-the-art results across various tasks \citep{devlin2019bert, liu2019roberta, lewis2020bart}. Their performance only continues to improve with the expansion of computational power, the availability of vast datasets, and the scaling of model architectures \citep{kaplan2020scaling, brown2020language, touvron2023llama}. But, despite their success, a significant portion of the world’s languages, particularly low-resource languages (LRLs), remain underrepresented in NLP research. These languages, numbering in the thousands, lack the necessary linguistic resources and data required for traditional statistical methods to be effectively applied \citep{singh2008low, cieri2016language, tsvetkov2017low}.

The challenges associated with LRLs are substantial and multifaceted. These languages often suffer from a lack of computational tools, limited digital presence, and insufficient educational infrastructure, which hinders their incorporation into modern NLP systems. While advancements in NLP for LRLs present immense potential, especially in regions such as Africa and India, where over 2.5 billion people speak these languages, the barriers are high. Addressing these challenges not only promises economic and cultural benefits but also plays a vital role in preserving linguistic heritage and improving societal outcomes, such as aiding in emergency response and enhancing educational and cultural exchanges \citep{tsvetkov2017low}.

\subsection{Knowledge Distillation}
Knowledge distillation (KD) refers to the process where a smaller model learns from a larger one, transferring knowledge from a teacher model to a student model to enhance the student’s performance. The central concept is that the student model emulates the teacher model, using the teacher's insights to achieve competitive or even superior results. A typical KD framework includes three main components: knowledge, a distillation algorithm, and the teacher-student model architecture. The key challenge is to effectively convey knowledge from the large teacher model to the smaller student model while ensuring the student retains or improves its performance \citep{bucilua2006model, ba2014do, hinton2015distilling, urban2017deep}.

Response-based knowledge distillation focuses on aligning the final output (logits) of the student model with the teacher's logits, typically using soft targets and temperature scaling to capture the "dark knowledge" from the teacher. This approach has been widely applied in tasks like image classification, but it is limited by only utilizing the output of the last layer, potentially missing intermediate-level supervision that could benefit deeper networks \citep{hinton2015distilling, ba2014do}. 

To address this, feature-based knowledge distillation extends the method by transferring feature maps from intermediate layers of the teacher model to the student model, enhancing representation learning. Techniques like attention maps and activation boundaries are used to match features between layers, but challenges persist in selecting the right layers and handling size differences between layers \citep{romero2015fitnets, zagoruyko2017paying}.
\\\\
In this work, we propose a novel hybrid distillation framework that integrates knowledge distillation and attention matching to improve multilingual model compression. Existing approaches often focus on either response-based distillation, which transfers knowledge through soft output distributions, or feature-based distillation, which aligns internal representations. Our method integrates both, enabling a more comprehensive transfer of knowledge from teacher to student models. Additionally, we introduce a highly compact multilingual student model with a significantly smaller hidden dimension, optimized for low-resource African languages. Our contributions can be summarized as follows:

\begin{itemize}
    \item \textbf{Hybrid Distillation Framework:} We propose a novel distillation approach that integrates knowledge distillation with attention matching, enabling the student model to learn both the output distribution and internal attention patterns of the teacher model.  
    \item \textbf{Compact Multilingual Model:} We introduce an extremely compact multilingual architecture with a hidden dimension of 256, significantly smaller than existing models (typically 768 or higher), while maintaining reasonable performance.  
    \item \textbf{Simplified Attention Matching:} Our mean-pooled attention matching mechanism effectively transfers knowledge between teacher and student models while reducing computational overhead compared to complex relation-based methods.  
    \item \textbf{Evaluation on African Languages:} We conduct a systematic evaluation of our hybrid distillation approach on five African languages—Kinyarwanda (\texttt{rw}), Swahili (\texttt{sw}), Hausa (\texttt{ha}), Igbo (\texttt{ig}), and Yoruba (\texttt{yo})—addressing gaps in model compression research for low-resource languages.  
    \item \textbf{Empirical Analysis:} We analyze trade-offs between model size, computational efficiency, and performance, demonstrating that effective knowledge transfer is possible despite substantial architectural differences between teacher and student models.  
\end{itemize}


\section{Related Work}
\citet{alabi2022adapting} proposed AfroXLMR, a multilingual pre-trained language model specifically adapted for African languages through multilingual adaptive fine-tuning (MAFT). Their approach has shown to enhance the performance of pre-trained models, such as XLM-R and AfriBERTa, on a variety of tasks for African languages by fine-tuning on monolingual texts from 17 high-resource African languages, alongside three widely spoken high-resource languages in Africa. One of the key innovations is the reduction of model size by removing tokens for non-African writing scripts from the embedding layer, which decreases the model size by approximately 50\%. The resulting model not only performs competitively with language-adaptive fine-tuning (LAFT) on individual languages but also improves the cross-lingual transfer \cite{thangaraj2024crosslingual} abilities of models like XLM-R, while requiring significantly less disk space. This makes the AfroXLMR approach more efficient for practical deployment on tasks such as Named Entity Recognition (NER), topic classification, and sentiment analysis, especially in low-resource African languages.

In MiniLM \citep{wang2020minilm}, deep self-attention distillation is used to compress pre-trained Transformers by transferring knowledge from the teacher model to the student model. MiniLMv2 \citep{wang2020minilmv2} builds on this by introducing multi-head self-attention relation distillation for task-agnostic compression, where attention relations are defined as the scaled dot-product between query, key, and value pairs within the self-attention module. Unlike previous methods, MiniLMv2 allows the student model to have a different number of attention heads than the teacher, which removes the constraint of matching attention head numbers. By concatenating queries from multiple attention heads and splitting them to match the desired number of relation heads, MiniLMv2 enables more fine-grained attention knowledge transfer, leading to a deeper mimicry of the teacher’s attention mechanisms. Moreover, MiniLMv2 examines layer selection beyond just the last layer, and finds that transferring knowledge from an upper-middle layer results in improved performance, especially for large models. Experimental results demonstrate that MiniLMv2, applied to both monolingual and multilingual pre-trained models, outperforms state-of-the-art methods, achieving better performance with fewer training examples and faster execution times.


\section{Methodology}

\begin{figure}[hbt]
  \includegraphics[width=\columnwidth]{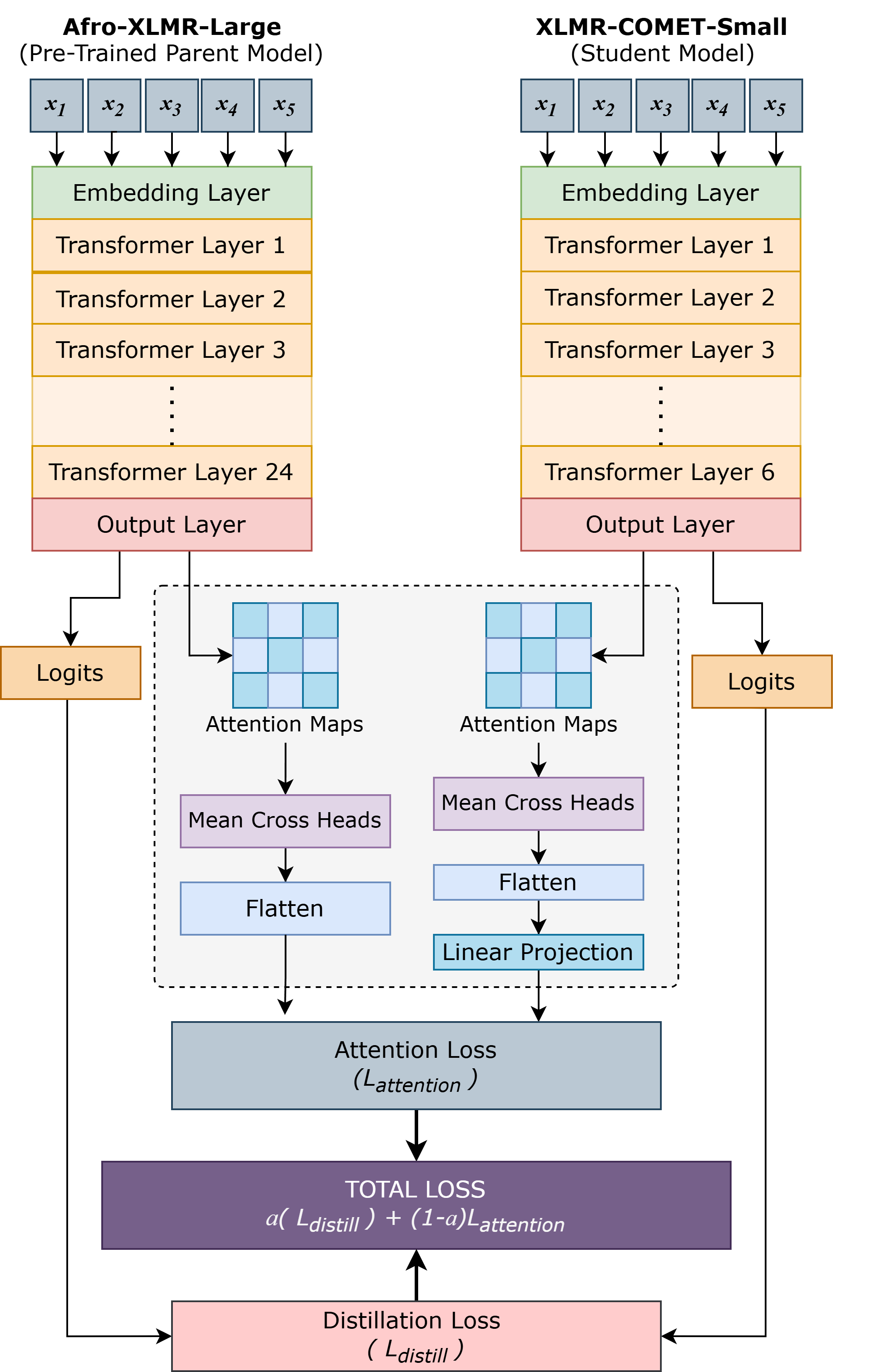}
  \caption{Proposed hybrid distillation framework.}
  \label{fig:framework}
\end{figure}

\subsection{Model Architecture}

\subsubsection{Teacher Model}
For our teacher model, we utilize the Afro-XLM-RoBERTa-Large (AfroXLMR-Large) architecture, a multilingual language model specifically pre-trained on on 17 African languages (Afrikaans, Amharic, Hausa, Igbo, Malagasy, Chichewa, Oromo, Naija, Kinyarwanda, Kirundi, Shona, Somali, Sesotho, Swahili, isiXhosa, Yoruba, and isiZulu) covering the major African language families and 3 high-resource languages (Arabic, French, and English). This model follows the standard transformer architecture with a hidden size of 1024, 16 attention heads, and 24 transformer layers (refer Table \ref{tab:model-config}). The model was pre-trained on a diverse corpus of African language texts, making it particularly suitable for our target languages.

\subsubsection{Student Model}
Our student model architecture is derived from XLM-RoBERTa-COMET-small, leveraging the mMiniLM-L12xH384 XLM-R model proposed by \citet{wang2020minilmv2}. However, we introduce significant modifications to further reduce the model size while maintaining performance (refer to Table \ref{tab:model-config}). The key architectural changes include:

1. \textbf{Hidden Size Reduction:} We reduce the hidden size from the original 384 to 256 dimensions, to further decreasing the model's parameter count. 

2. \textbf{Intermediate Layer:} To maintain architectural balance with the reduced hidden size, we adjust the intermediate layer size to 1024, compared to the original 1536 dimensions.

3. \textbf{Attention Head:} We configure the model with 8 attention heads, ensuring that each head operates on 32-dimensional key, query, and value vectors (256/8 = 32), maintaining the standard practice of having head dimensions that are factors of the hidden size.

This architectural configuration results in a substantially more compact model while preserving the essential multi-head attention mechanism necessary for capturing complex linguistic patterns. The reduced dimensionality is compensated for through our attention matching mechanism, which enables the student model to learn effective representations despite its smaller size.

\begin{table*}[hbt!]
    \centering
    \begin{tabular}{l|c|c|c}
        \hline
        \textbf{Attribute} & \textbf{AfroXLMR-Large} & \textbf{XLMR-Comet-Small} & \textbf{AfroXLMR-Comet (proposed)} \\
        \hline
        Hidden Size & 1024 & 384 & 256 \\
        Attention Heads & 16 & 12 & 8 \\
        Hidden Layers & 24 & 6 & 6 \\
        Intermediate Size & 4096 & 1536 & 1024 \\
        Parameter Count & 559,890,432 & 106,993,920 & 68,937,216\\
        \hline
    \end{tabular}
    \caption{Comparison of model configurations and parameter counts, highlighting the reduction in parameters for the Student model.}
    \label{tab:model-config}
\end{table*}

\subsection{Dataset Preparation}
In this work, we focus on a multilingual dataset--the MADLAD-400 dataset \citep{kudugunta2023madlad400}, a manually audited, general domain 3T token monolingual dataset based on CommonCrawl, spanning 419 languages. We employ a multilingual subset of the  dataset that represents African languages, specifically Kinyarwanda (rw), Swahili (sw), Hausa (ha), Igbo (ig), and Yoruba (yo). The dataset is then split into 80\% for training and 20\% for evaluation. All sequences are tokenized to a maximum length of 128 tokens.
\begin{table}[h!]
    \centering
    \begin{tabular}{l|cc}
        \hline
        \textbf{Lang.} & \textbf{Sequences} & \textbf{Sequences Used} \\
        \hline
        rw & 226,466 & 226,466 \\
        sw & 537,847 & 250,000 \\
        ha & 173,485 & 173,485 \\
        ig & 54,410 & 54,410 \\
        yo & 52,067 & 52,067 \\
        \hline
    \end{tabular}
    \caption{Total and used sequences from the Madlad-400 dataset for selected languages}
    \label{tab:madlad-sequence-usage}
\end{table}

\subsection{Attention Matching Mechanism}

\begin{algorithm}[hbt!]
\caption{Multilingual Knowledge Distillation}
\label{alg:distillation}
\begin{algorithmic}[1]
\Require Student model $S$, Teacher model $T$, Temperature $\tau$, Weight factor $\alpha$, Projection layer $P$
\Require Training dataset $\mathcal{D}$, Loss function $\mathcal{L}_{\text{MSE}}$
\For{each batch $(X, Y) \in \mathcal{D}$}
    \State \textbf{Forward Pass:}
    \State $Z_T \gets T(X)$ \Comment{Teacher logits (no gradient)}
    \State $Z_S \gets S(X)$ \Comment{Student logits}
    
    \State \textbf{Distillation Loss:}
    \State $P_T \gets \text{softmax}(Z_T / \tau)$
    \State $P_S \gets \text{softmax}(Z_S / \tau)$
    \State $\mathcal{L}_{\text{distill}} \gets -\sum P_T \log (P_S + \epsilon)$
    
    \If{attention available in $S$ and $T$}
        \State \textbf{Attention Loss:}
        \State $A_T \gets \text{mean}(T.\text{attentions}[-1], \text{dim}=1)$
        \State $A_S \gets \text{mean}(S.\text{attentions}[-1], \text{dim}=1)$
        \State $A_T \gets \text{flatten}(A_T)$, \quad $A_S \gets \text{flatten}(A_S)$
        \State $A_S' \gets P(A_S)$ \Comment{Project student attention}
        \State $\mathcal{L}_{\text{attn}} \gets \mathcal{L}_{\text{MSE}}(A_S', A_T)$
        \State $\mathcal{L} \gets \alpha \mathcal{L}_{\text{distill}} + (1 - \alpha) \mathcal{L}_{\text{attn}}$
    \Else
        \State $\mathcal{L} \gets \mathcal{L}_{\text{distill}}$
    \EndIf
    
    \State \textbf{Backward Pass:}
    \State Update $S$ using $\nabla \mathcal{L}$
\EndFor
\end{algorithmic}
\end{algorithm}

Our work implements a simplified attention matching approach that focuses on the aggregate attention patterns rather than individual head-level interactions. While previous work, such as MiniLMv2 \citep{wang2020minilmv2}, addresses head count mismatches through relation heads and KL-divergence, we propose a more streamlined approach that captures the overall attention behavior.

We extract the attention matrices from the final layer of both teacher and student models. To manage the dimensional differences between the models, we first compute the mean across attention heads, producing a single attention map per sequence. This averaging operation reduces the attention tensors from shape \textit{[batch\_size, num\_heads, sequence\_length, sequence\_length]} to \textit{[batch\_size, sequence\_length, sequence\_length]}, effectively capturing the aggregate attention patterns across all heads.

The attention matrices are then flattened to vectors of shape \textit{[batch\_size, sequence\_length} x \textit{sequence\_length]}. To address the remaining dimensional differences between teacher and student models, we employ a learnable linear projection layer that maps the flattened student attention patterns to the teacher's dimensional space. The projection operation can be formalized as:
\begin{equation}    
`   A_{\text{student\_proj}} = W \cdot A_{\text{student\_flat}}
\end{equation}
\newpage
 where:
\begin{itemize}
    \setlength{\itemsep}{0pt} 
    \setlength{\parskip}{0pt}
    \item $A_{\text{student\_flat}}$ is the flattened student attention matrix.
    \item $W$ is the learned projection matrix.
    \item $A_{\text{student\_proj}}$ is the projected student attention pattern.
\end{itemize}

The training objective for attention matching is computed using Mean Squared Error (MSE) loss between the projected student attention and the flattened teacher attention:
\begin{equation}
    L_{\text{attention}} = \text{MSE}(A_{\text{student\_proj}}, A_{\text{teacher\_flat}})
\end{equation}
where:
\begin{itemize}
    \setlength{\itemsep}{0pt}  
    \setlength{\parskip}{0pt}  
    \item $L_{\text{attention}}$ is the attention loss function.  
    \item $\text{MSE}$ is the Mean Squared Error function.  
    \item $A_{\text{student\_proj}}$ is the projected student attention pattern.  
    \item $A_{\text{teacher\_flat}}$ is the flattened teacher attention matrix.  
\end{itemize}

This approach, while simpler than the relation-based methods, proves effective in practice. By focusing on the aggregate attention patterns rather than individual head interactions, we reduce computational complexity while still capturing the essential aspects of the teacher's attention mechanism. The learned projection layer allows the student to adapt its attention patterns to match the teacher's higher-dimensional space, facilitating effective knowledge transfer despite the architectural differences between the models.

Our final loss function combines this attention matching loss with traditional knowledge distillation:
\begin{equation}
    L_{\text{total}} = \alpha \cdot L_{\text{distill}} + (1-\alpha) \cdot L_{\text{attention}}
\end{equation}
where:
\begin{itemize}
    \item $\alpha = 0.5$ balances between the distillation and attention matching objectives.
\end{itemize}

\subsection{Training Process}
To facilitate effective knowledge transfer between these architecturally different models, we employ a two-stage training process (refer Figure \ref{fig:framework}). First, we initialize the student model with the reduced configuration parameters. Then, we apply our combined distillation approach, using both soft target probabilities and attention matching to transfer knowledge from the teacher to the student model. 

Instead of task-specific distillation, we follow a task-agnostic knowledge distillation framework, where the student model is trained to generalize knowledge from the teacher across multiple tasks, rather than being optimized for a single downstream task. By distilling general linguistic representations instead of task-specific knowledge, our model remains adaptable to various NLP applications without requiring retraining for different use cases.

For soft target distillation, we use a temperature parameter T=2.0 to create smoothed probability distributions from both models' logits, then measure their difference using Kullback-Leibler (KL) divergence. This smoothing reveals the teacher's underlying knowledge through relative probabilities across all classes, while KL divergence effectively captures how the student's probability distribution diverges from the teacher's desired distribution. The soft probability for class \textit{i} is calculated as:
\begin{equation}
    p_i = \frac{\exp(z_i/T)}{\sum_{j=1}^N \exp(z_j/T)}
\end{equation}
where:
\begin{itemize}
    \setlength{\itemsep}{0pt}  
    \setlength{\parskip}{0pt} 
    \item $z_i$ is the logit (raw score) for class i
    \item T is the temperature parameter (typically T>1)
    \item N is the total number of classes
    \item $p_i$ is the resulting soft probability for class i
\end{itemize}

\begin{table}[hbt!]
    \centering
    \begin{tabular}{l|l}
        \hline
        \textbf{Parameter} & \textbf{Value} \\  
        \hline
        Max Sequence Length & 128 \\
        Padding & True \\
        Truncation & True \\
        Evaluation Strategy & Epoch \\  
        Logging Steps & 50 \\  
        Learning Rate & \(5 \times 10^{-5}\) \\  
        Number of Training Epochs & 15 \\  
        Per Device Train Batch Size & 8 \\  
        Gradient Accumulation Steps & 2 \\  
        Mixed Precision (FP16) & True \\  
        Save Strategy & Epoch \\  
        Save Total Limit & 3 \\  
        Load Best Model at End & True \\  
        Early Stopping Patience & 3 \\  
        Early Stopping Threshold & 0.01 \\  
        \hline
    \end{tabular}
    \caption{Training parameters and values used for model distillation.}
    \label{tab:training-params}
\end{table}

\begin{figure}[hbt!]
  \includegraphics[width=\columnwidth]{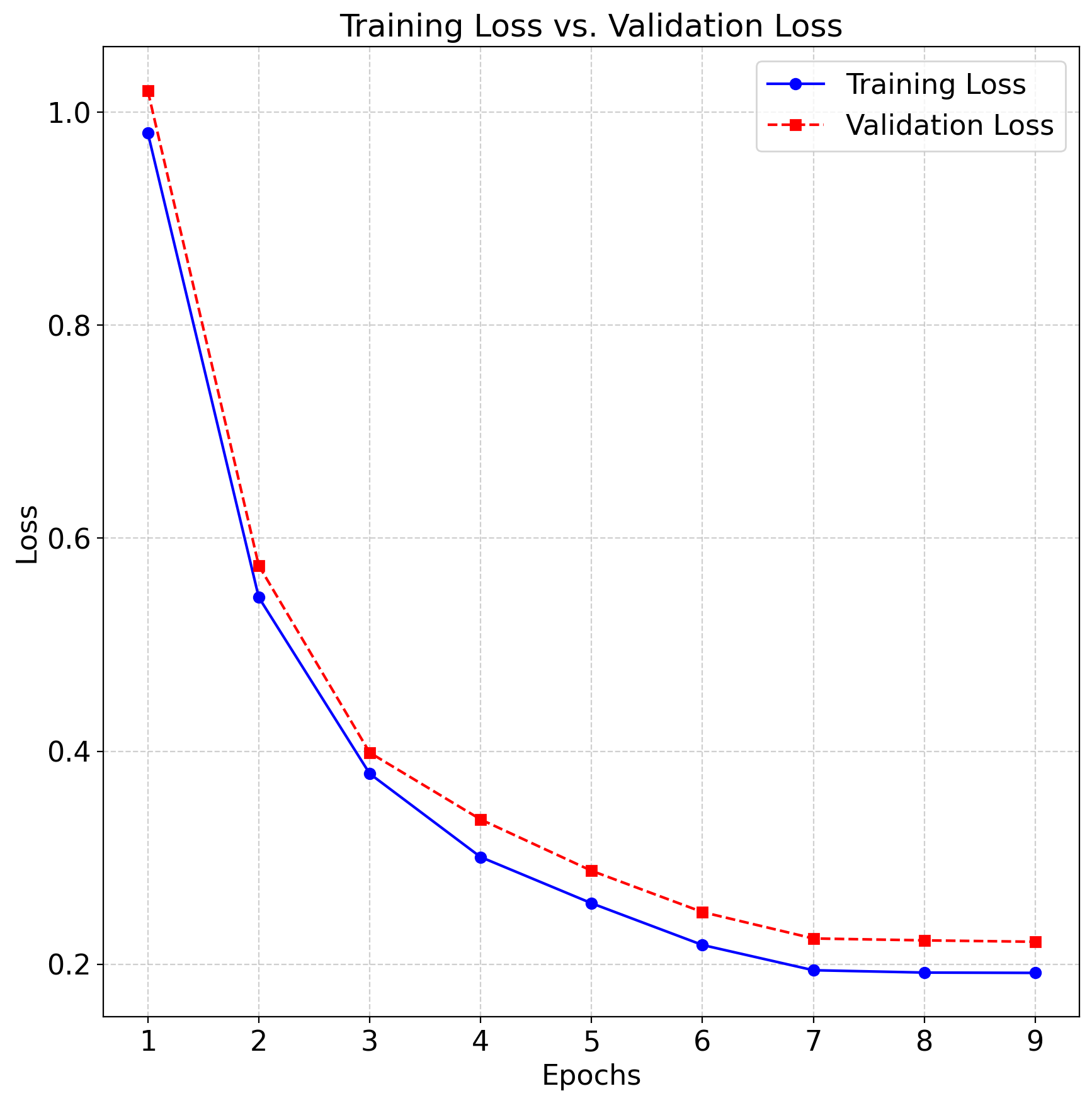}
  \caption{Training and validation loss curves showing model convergence over 9 epochs. Training terminated via early stopping to prevent overfitting.}
  \label{fig:loss-graph}
\end{figure}

The training process is optimized using AdamW optimizer with a learning rate of \textit{5e-5}. We implement early stopping with a patience of 3 epochs and a threshold of 0.01 to prevent overfitting, ensuring the student model achieves optimal performance without unnecessary computation. Refer to Table \ref{tab:training-params} for other training parameters for the model distillation process, Figure \ref{fig:loss-graph} for the training and validation loss curves, and to Algorithm \ref{alg:distillation} for the training process' algorithm.

\section{Experiments}
We evaluate our distilled student model--AfroXLMR-Comet--using the AfriSenti dataset from the AfriSenti-SemEval\citep{muhammad-etal-2023-afrisenti, muhammadSemEval2023, muhammad-etal-2022-naijasenti, yimametalcoling2020} Shared Task 12, which covers 17 African languages: Hausa, Yoruba, Igbo, Nigerian Pidgin, Amharic, Tigrinya, Oromo, Swahili, Algerian Arabic, Kinyarwanda, Twi, Mozambican Portuguese, Moroccan Arabic, Fongbe, Lingala, Kamba, and Luganda. For our comparative analysis with AfroXLMR models, we focus on five major languages: \textit{Kinyarwanda, Swahili, Hausa, Igbo,} and \textit{Yoruba}. The dataset consists of manually annotated tweets categorized into positive, negative, and neutral sentiments (refer Table \ref{tab:dataset-fields}. We maintain the original data splits provided by the task organizers for consistent comparison with benchmark models (refer Table \ref{tab:afrisent-samples}). Each model was fine-tuned for 5 epochs at a learning rate of \textit{2e-5} on each language dataset provided by the task and then retrained for Task A. Our benchmark results are presented in Table \ref{tab:accuracy-comparison}.

\begin{table}[hbt!]
    \centering
    \begin{tabular}{l|l}
        \hline
        \textbf{Field} & \textbf{Description} \\  
        \hline
        ID     & Alpha-Numeric Serial Numbers \\  
        Tweet  & Tweet Content \\  
        Label  & Tweet Sentiment Label \\  
        \hline
    \end{tabular}
    \caption{Field descriptions of the dataset.}
    \label{tab:dataset-fields}
\end{table}

\begin{table}[hbt!]
    \centering
    \begin{tabular}{l|r r}
        \hline
        \textbf{Language} & \textbf{Train} & \textbf{Test} \\
        \hline
        kin & 3,302 & 1,026 \\
        swa & 1,810 & 748 \\
        hau & 14,172 & 5,303 \\
        igo & 10,192 & 3,682 \\
        yor & 8,522 & 4,515 \\
        \hline
    \end{tabular}
    \caption{Number of training and testing samples for each language in the AfriSenti dataset}
    \label{tab:afrisent-samples}
\end{table}

\begin{table*}[hbt!]
  \centering
  \renewcommand{\arraystretch}{1.2}
  \setlength{\tabcolsep}{10pt} 

  \begin{tabular}{l|ccccc|c}
    \hline
    \textbf{Model} & \textbf{kin} & \textbf{swa} & \textbf{hau} & \textbf{igo} & \textbf{yor} & \textbf{avg} \\
    \hline
    AfroXLMR-Large (Parent)  & 68.91  & 64.10  & 79.30  & 79.60  & 74.20  & 73.22  \\
    AfroXLMR-Base            & 62.54  & 60.81  & 78.27  & 78.63  & 70.25  & 70.10  \\
    AfroXLMR-Mini            & 60.23  & 61.88  & 74.29  & 75.20  & 63.10  & 66.94
  \\
    \rowcolor[gray]{0.9} AfroXLMR-Comet (Student) & 58.94  & 55.30  & 71.89  & 73.82  & 66.39  & 65.28
  \\
    \hline
  \end{tabular}
  
  \caption{\label{tab:accuracy-comparison}
    Sentiment classification model comparison on AfriSenti-SemEval, showing F1 evaluation on test sets after 5 epochs. The last column represents the average F1 score across all languages.
  }
\end{table*}

\begin{table*}[hbt!]
  \centering
  \renewcommand{\arraystretch}{1.2}
  \setlength{\tabcolsep}{8pt}
  
  \begin{tabular}{l|ccc}
    \hline
    \textbf{Model}           & \textbf{Parameter Count}      & \textbf{Inference Time (ms)} & \textbf{Size (MB)} \\
    \hline
    AfroXLMR-Large (Parent)  & 559,890,432  & 293.9  & 2135.86  \\
    AfroXLMR-Base            & 278,043,648  & 100.5  & 1060.67  \\
    AfroXLMR-Mini            & 117,640,704  & 30.2   & 448.79   \\
    \rowcolor[gray]{0.9} AfroXLMR-Comet (Student) & 68,937,216   & 14.0   & 262.99   \\
    \hline
  \end{tabular}
  
  \caption{\label{model-comparison} 
    Comparison of AfroXLMR-Large, AfroXLMR-Base, AfroXLMR-Mini, and our distilled model in terms of parameter count, inference time, and model size. The proposed distilled model is significantly smaller and the fastest among the models.
  }
  \label{tab:afro-fam-comp}
\end{table*}

Further in Table \ref{tab:afro-fam-comp}, the parameter count, inference time, and model size comparison highlight the efficiency of the proposed AfroXLMR-Comet model. AfroXLMR-Large, with the highest parameter count (559.9M), also has the largest model size (2.09 GB) and the slowest inference time (293.9 milliseconds) due to its significant computational requirements. AfroXLMR-Base (278M parameters, 1.04 GB) provides a better balance, achieving an inference time of 100.5 milliseconds, while AfroXLMR-Mini (117.6M parameters, 448.79 MB) further reduces computational load with a faster inference time of 30.2 milliseconds. Our AfroXLMR-Comet model, with just 68.9M parameters and a compact size of 262.99 MB, achieves the fastest inference time of 14.0 milliseconds, making it the most efficient among all models.


\section{Results and Discussion}
The distilled student model--AfroXLMR-Comet--achieved a substantial parameter reduction of 87.69\%, compressing the teacher model's 559,890,432 parameters down to 68,937,216. This reduction in size results in a trade-off in performance, as the student model achieves an average F1 score of 65.28\% compared to Afro-XLMR-Large’s 73.22\% on the AfriSenti-SemEval sentiment classification benchmark. Despite this drop in accuracy, the student model significantly reduces inference latency and memory footprint, making it particularly suitable for deployment in resource-constrained environments. Additionally, the student model’s compact size of 262.99 MB (compared to 2.09 GB for Afro-XLMR-Large) highlights its efficiency in handling low-resource languages.

Notably, the distilled student model's performance is very close to Afro-XLMR-Mini, which achieves an F1 score of 66.94\% but is almost twice the size at 448.79 MB, further showcasing that the student model balances efficiency and performance, achieving comparable results with a significantly smaller memory footprint. The simplification of attention mechanisms further emphasizes the practicality of this approach, ensuring efficient knowledge transfer while maintaining competitive performance across multiple African languages.


\section{Conclusion}
In this work, we introduced a hybrid distillation framework for the efficient compression of pre-trained multilingual transformer models, specifically focusing on African languages. By integrating task-agnostic knowledge distillation, attention matching, and adaptive learning mechanisms, we successfully distilled AfroXLMR-Large into a lightweight yet effective student model. Our approach achieves a significant reduction in computational and memory requirements while maintaining competitive performance, demonstrating a viable trade-off between efficiency and accuracy.

Looking ahead, future work will explore extending this methodology to other language families and investigating domain-specific fine-tuning to further enhance the adaptability of compressed multilingual models. Additionally, we aim to refine attention transfer mechanisms to improve performance retention in extreme compression settings. Our findings highlight the transformative potential of efficient model compression in democratizing access to NLP for low-resource languages, ensuring that cutting-edge language technologies become more accessible to linguistic communities worldwide.


\section{Limitations}
While our framework shows promising results, several important limitations should be acknowledged:
\begin{itemize}
    \item \textbf{Projection Layer Computation:} The projection layer matching between teacher and student attention matrices introduces additional computational overhead. For a sequence length of 128, the projection layer requires maintaining and computing gradients for a large parameter matrix of size (128²) × (128²), which can be memory-intensive during training. This limitation becomes more pronounced when dealing with longer sequences or larger batch sizes, leading to memory allocation challenges during both fine-tuning and inference. This limitation, while less severe than head-level matching approaches, still impacts the model's deployability in resource-constrained environments.   
    
    \item \textbf{Data and Resource Constraints:} A fundamental challenge in our work is the scarcity of high-quality dataset samples for many low-resource languages, which inherently limits the extent of multilingual generalization. Due to computational constraints, our experimental validation was limited to five languages.
    
    \item \textbf{Temperature Sensitivity:} The distillation process relies heavily on the temperature parameter $\tau$ in the softmax computations. While we use a fixed temperature of 2.0, the optimal temperature may vary across different languages and tasks. Our approach lacks adaptive temperature scaling that could potentially improve knowledge transfer for specific language combinations.
\end{itemize}




\bibliography{main}

\begin{thebibliography}{24}
\providecommand{\natexlab}[1]{#1}

\bibitem[{Alabi et~al.(2022)Alabi, Adelani, Mosbach, and Klakow}]{alabi2022adapting}
Jesujoba~O. Alabi, David~Ifeoluwa Adelani, Marius Mosbach, and Dietrich Klakow. 2022.
\newblock \href {https://arxiv.org/abs/2204.06487} {Adapting pre-trained language models to african languages via multilingual adaptive fine-tuning}.
\newblock \emph{Preprint}, arXiv:2204.06487.

\bibitem[{Ba and Caruana(2014)}]{ba2014do}
Jimmy Ba and Rich Caruana. 2014.
\newblock \href {https://proceedings.neurips.cc/paper_files/paper/2014/file/ea8fcd92d59581717e06eb187f10666d-Paper.pdf} {Do deep nets really need to be deep?}
\newblock In \emph{Advances in Neural Information Processing Systems}, volume~27. Curran Associates, Inc.

\bibitem[{Brown et~al.(2020)Brown, Mann, Ryder, Subbiah, Kaplan, Dhariwal, Neelakantan, Shyam, Sastry, Askell, Agarwal, Herbert-Voss, Krueger, Henighan, Child, Ramesh, Ziegler, Wu, Winter, Hesse, Chen, Sigler, Litwin, Gray, Chess, Clark, Berner, McCandlish, Radford, Sutskever, and Amodei}]{brown2020language}
Tom Brown, Benjamin Mann, Nick Ryder, Melanie Subbiah, Jared~D Kaplan, Prafulla Dhariwal, Arvind Neelakantan, Pranav Shyam, Girish Sastry, Amanda Askell, Sandhini Agarwal, Ariel Herbert-Voss, Gretchen Krueger, Tom Henighan, Rewon Child, Aditya Ramesh, Daniel Ziegler, Jeffrey Wu, Clemens Winter, Chris Hesse, Mark Chen, Eric Sigler, Mateusz Litwin, Scott Gray, Benjamin Chess, Jack Clark, Christopher Berner, Sam McCandlish, Alec Radford, Ilya Sutskever, and Dario Amodei. 2020.
\newblock \href {https://proceedings.neurips.cc/paper_files/paper/2020/file/1457c0d6bfcb4967418bfb8ac142f64a-Paper.pdf} {Language models are few-shot learners}.
\newblock In \emph{Advances in Neural Information Processing Systems}, volume~33, pages 1877--1901. Curran Associates, Inc.

\bibitem[{Buciluǎ et~al.(2006)Buciluǎ, Caruana, and Niculescu-Mizil}]{bucilua2006model}
Cristian Buciluǎ, Rich Caruana, and Alexandru Niculescu-Mizil. 2006.
\newblock Model compression.
\newblock In \emph{Proceedings of the 12th ACM SIGKDD international conference on Knowledge discovery and data mining}, pages 535--541.

\bibitem[{Cieri et~al.(2016)Cieri, Maxwell, Strassel, and Tracey}]{cieri2016language}
Christopher Cieri, Mike Maxwell, Stephanie Strassel, and Jennifer Tracey. 2016.
\newblock \href {https://aclanthology.org/L16-1720/} {Selection criteria for low resource language programs}.
\newblock In \emph{Proceedings of the Tenth International Conference on Language Resources and Evaluation ({LREC}`16)}, pages 4543--4549, Portoro{\v{z}}, Slovenia. European Language Resources Association (ELRA).

\bibitem[{Devlin(2018)}]{devlin2019bert}
Jacob Devlin. 2018.
\newblock Bert: Pre-training of deep bidirectional transformers for language understanding.
\newblock \emph{arXiv preprint arXiv:1810.04805}.

\bibitem[{Hinton(2015)}]{hinton2015distilling}
Geoffrey Hinton. 2015.
\newblock Distilling the knowledge in a neural network.
\newblock \emph{arXiv preprint arXiv:1503.02531}.

\bibitem[{Kaplan et~al.(2020)Kaplan, McCandlish, Henighan, Brown, Chess, Child, Gray, Radford, Wu, and Amodei}]{kaplan2020scaling}
Jared Kaplan, Sam McCandlish, Tom Henighan, Tom~B. Brown, Benjamin Chess, Rewon Child, Scott Gray, Alec Radford, Jeffrey Wu, and Dario Amodei. 2020.
\newblock \href {https://arxiv.org/abs/2001.08361} {Scaling laws for neural language models}.
\newblock \emph{Preprint}, arXiv:2001.08361.

\bibitem[{Kudugunta et~al.(2023)Kudugunta, Caswell, Zhang, Garcia, Xin, Kusupati, Stella, Bapna, and Firat}]{kudugunta2023madlad400}
Sneha Kudugunta, Isaac Caswell, Biao Zhang, Xavier Garcia, Derrick Xin, Aditya Kusupati, Romi Stella, Ankur Bapna, and Orhan Firat. 2023.
\newblock \href {https://proceedings.neurips.cc/paper_files/paper/2023/file/d49042a5d49818711c401d34172f9900-Paper-Datasets_and_Benchmarks.pdf} {Madlad-400: A multilingual and document-level large audited dataset}.
\newblock In \emph{Advances in Neural Information Processing Systems}, volume~36, pages 67284--67296. Curran Associates, Inc.

\bibitem[{Lewis(2019)}]{lewis2020bart}
Mike Lewis. 2019.
\newblock Bart: Denoising sequence-to-sequence pre-training for natural language generation, translation, and comprehension.
\newblock \emph{arXiv preprint arXiv:1910.13461}.

\bibitem[{Liu et~al.(2019)Liu, Ott, Goyal, Du, Joshi, Chen, Levy, Lewis, Zettlemoyer, and Stoyanov}]{liu2019roberta}
Yinhan Liu, Myle Ott, Naman Goyal, Jingfei Du, Mandar Joshi, Danqi Chen, Omer Levy, Mike Lewis, Luke Zettlemoyer, and Veselin Stoyanov. 2019.
\newblock \href {https://arxiv.org/abs/1907.11692} {Roberta: A robustly optimized bert pretraining approach}.
\newblock \emph{Preprint}, arXiv:1907.11692.

\bibitem[{Muhammad et~al.(2023{\natexlab{a}})Muhammad, Abdulmumin, Ayele, Ousidhoum, Adelani, Yimam, Ahmad, Beloucif, Mohammad, Ruder, Hourrane, Jorge, Brazdil, Ali, David, Osei, Shehu-Bello, Lawan, Gwadabe, Rutunda, Belay, Messelle, Balcha, Chala, Gebremichael, Opoku, and Arthur}]{muhammad-etal-2023-afrisenti}
Shamsuddeen Muhammad, Idris Abdulmumin, Abinew Ayele, Nedjma Ousidhoum, David Adelani, Seid Yimam, Ibrahim Ahmad, Meriem Beloucif, Saif Mohammad, Sebastian Ruder, Oumaima Hourrane, Alipio Jorge, Pavel Brazdil, Felermino Ali, Davis David, Salomey Osei, Bello Shehu-Bello, Falalu Lawan, Tajuddeen Gwadabe, Samuel Rutunda, Tadesse Belay, Wendimu Messelle, Hailu Balcha, Sisay Chala, Hagos Gebremichael, Bernard Opoku, and Stephen Arthur. 2023{\natexlab{a}}.
\newblock \href {https://doi.org/10.18653/v1/2023.emnlp-main.862} {{A}fri{S}enti: A {T}witter sentiment analysis benchmark for {A}frican languages}.
\newblock In \emph{Proceedings of the 2023 Conference on Empirical Methods in Natural Language Processing}, pages 13968--13981, Singapore. Association for Computational Linguistics.

\bibitem[{Muhammad et~al.(2023{\natexlab{b}})Muhammad, Abdulmumin, Yimam, Adelani, Ahmad, Ousidhoum, Ayele, Mohammad, Beloucif, and Ruder}]{muhammadSemEval2023}
Shamsuddeen~Hassan Muhammad, Idris Abdulmumin, Seid~Muhie Yimam, David~Ifeoluwa Adelani, Ibrahim~Sa'id Ahmad, Nedjma Ousidhoum, Abinew~Ali Ayele, Saif~M. Mohammad, Meriem Beloucif, and Sebastian Ruder. 2023{\natexlab{b}}.
\newblock {SemEval-2023 Task 12: Sentiment Analysis for African Languages (AfriSenti-SemEval)}.
\newblock In \emph{Proceedings of the 17th {{International Workshop}} on {{Semantic Evaluation}} ({{SemEval-2023}})}. {Association for Computational Linguistics}.

\bibitem[{Muhammad et~al.(2022)Muhammad, Adelani, Ruder, Ahmad, Abdulmumin, Bello, Choudhury, Emezue, Abdullahi, Aremu, orge, and Brazdil}]{muhammad-etal-2022-naijasenti}
Shamsuddeen~Hassan Muhammad, David~Ifeoluwa Adelani, Sebastian Ruder, Ibrahim~Sa{'}id Ahmad, Idris Abdulmumin, Bello~Shehu Bello, Monojit Choudhury, Chris~Chinenye Emezue, Saheed~Salahudeen Abdullahi, Anuoluwapo Aremu, Al{\'\i}pio orge, and Pavel Brazdil. 2022.
\newblock \href {https://aclanthology.org/2022.lrec-1.63} {{N}aija{S}enti: A {N}igerian {T}witter sentiment corpus for multilingual sentiment analysis}.
\newblock In \emph{Proceedings of the Thirteenth Language Resources and Evaluation Conference}, pages 590--602, Marseille, France. European Language Resources Association.

\bibitem[{Romero et~al.(2015)Romero, Ballas, Kahou, Chassang, Gatta, and Bengio}]{romero2015fitnets}
Adriana Romero, Nicolas Ballas, Samira~Ebrahimi Kahou, Antoine Chassang, Carlo Gatta, and Yoshua Bengio. 2015.
\newblock \href {https://arxiv.org/abs/1412.6550} {Fitnets: Hints for thin deep nets}.
\newblock \emph{Preprint}, arXiv:1412.6550.

\bibitem[{Singh(2008)}]{singh2008low}
Anil~Kumar Singh. 2008.
\newblock Natural language processing for less privileged languages: Where do we come from? where are we going?
\newblock In \emph{Proceedings of the IJCNLP-08 Workshop on NLP for Less Privileged Languages}.

\bibitem[{Thangaraj et~al.(2024)Thangaraj, Chenat, Walia, and Marivate}]{thangaraj2024crosslingual}
Harish Thangaraj, Ananya Chenat, Jaskaran~Singh Walia, and Vukosi Marivate. 2024.
\newblock \href {https://arxiv.org/abs/2409.10965} {Cross-lingual transfer of multilingual models on low resource african languages}.
\newblock \emph{Preprint}, arXiv:2409.10965.

\bibitem[{Touvron et~al.(2023)Touvron, Lavril, Izacard, Martinet, Lachaux, Lacroix, Rozière, Goyal, Hambro, Azhar, Rodriguez, Joulin, Grave, and Lample}]{touvron2023llama}
Hugo Touvron, Thibaut Lavril, Gautier Izacard, Xavier Martinet, Marie-Anne Lachaux, Timothée Lacroix, Baptiste Rozière, Naman Goyal, Eric Hambro, Faisal Azhar, Aurelien Rodriguez, Armand Joulin, Edouard Grave, and Guillaume Lample. 2023.
\newblock \href {https://arxiv.org/abs/2302.13971} {Llama: Open and efficient foundation language models}.
\newblock \emph{Preprint}, arXiv:2302.13971.

\bibitem[{Tsvetkov(2017)}]{tsvetkov2017low}
Yulia Tsvetkov. 2017.
\newblock Opportunities and challenges in working with low-resource languages.
\newblock \emph{Slides Part-1}, page~39.

\bibitem[{Urban et~al.(2017)Urban, Geras, Kahou, Aslan, Wang, Caruana, Mohamed, Philipose, and Richardson}]{urban2017deep}
Gregor Urban, Krzysztof~J. Geras, Samira~Ebrahimi Kahou, Ozlem Aslan, Shengjie Wang, Rich Caruana, Abdelrahman Mohamed, Matthai Philipose, and Matt Richardson. 2017.
\newblock \href {https://arxiv.org/abs/1603.05691} {Do deep convolutional nets really need to be deep and convolutional?}
\newblock \emph{Preprint}, arXiv:1603.05691.

\bibitem[{Wang et~al.(2021)Wang, Bao, Huang, Dong, and Wei}]{wang2020minilmv2}
Wenhui Wang, Hangbo Bao, Shaohan Huang, Li~Dong, and Furu Wei. 2021.
\newblock \href {https://arxiv.org/abs/2012.15828} {Minilmv2: Multi-head self-attention relation distillation for compressing pretrained transformers}.
\newblock \emph{Preprint}, arXiv:2012.15828.

\bibitem[{Wang et~al.(2020)Wang, Wei, Dong, Bao, Yang, and Zhou}]{wang2020minilm}
Wenhui Wang, Furu Wei, Li~Dong, Hangbo Bao, Nan Yang, and Ming Zhou. 2020.
\newblock \href {https://proceedings.neurips.cc/paper_files/paper/2020/file/3f5ee243547dee91fbd053c1c4a845aa-Paper.pdf} {Minilm: Deep self-attention distillation for task-agnostic compression of pre-trained transformers}.
\newblock In \emph{Advances in Neural Information Processing Systems}, volume~33, pages 5776--5788. Curran Associates, Inc.

\bibitem[{Yimam et~al.(2020)Yimam, Alemayehu, Ayele, and Biemann}]{yimametalcoling2020}
Seid~Muhie Yimam, Hizkiel~Mitiku Alemayehu, Abinew Ayele, and Chris Biemann. 2020.
\newblock Exploring {A}mharic sentiment analysis from social media texts: Building annotation tools and classification models.
\newblock In \emph{Proceedings of the 28th International Conference on Computational Linguistics}, pages 1048--1060, Barcelona, Spain (Online).

\bibitem[{Zagoruyko and Komodakis(2017)}]{zagoruyko2017paying}
Sergey Zagoruyko and Nikos Komodakis. 2017.
\newblock \href {https://arxiv.org/abs/1612.03928} {Paying more attention to attention: Improving the performance of convolutional neural networks via attention transfer}.
\newblock \emph{Preprint}, arXiv:1612.03928.

\end{thebibliography}

\end{document}